\documentclass[letterpaper, 10 pt, conference]{ieeeconf}  

\IEEEoverridecommandlockouts                              

\usepackage{adjustbox}
\usepackage[ruled, vlined]{algorithm2e}
\usepackage{amsmath}
\usepackage{amssymb}
\usepackage{array}  
\usepackage{authblk}
\usepackage[accsupp]{axessibility}  
\usepackage{booktabs}
\usepackage{capt-of}
\makeatletter
\let\NAT@parse\undefined
\makeatother
\usepackage{cite}
\usepackage[font={small}]{caption}
\usepackage{colortbl}
\usepackage{epsfig}
\usepackage{inconsolata}  
\usepackage{graphicx}
\usepackage{listings}
\usepackage{multicol}
\usepackage{multirow}
\usepackage{paralist}
\usepackage{pifont}
\usepackage{ragged2e} 
\usepackage{times}
\usepackage{tabularx}
\usepackage{xcolor}
\usepackage{svg}

\definecolor{flodarkpurple}{rgb}{0.288,0.1196,0.7}

\definecolor{amber}{rgb}{1.0, 0.75, 0.0}

\definecolor{flodarkpurple}{rgb}{0.288,0.1196,0.7}

\definecolor{amber}{rgb}{1.0, 0.75, 0.0}

\usepackage[backref=page,breaklinks,colorlinks,bookmarks,citecolor=flodarkpurple]{hyperref}
\usepackage{multirow}

\newcommand{\webpage}{https://concept-agent.github.io/}

\title{\Large \bf
ConceptAgent: LLM-Driven Precondition Grounding and Tree Search for Robust Task Planning and Execution \\
}

\author{
Corban Rivera$^{\dagger*1,2}$,
Grayson Byrd$^{*1,2}$,
William Paul$^{*1,2}$,
Tyler Feldman$^{1}$,
Meghan Booker$^{1}$,
Emma Holmes$^{1}$, \\
David Handelman$^{1}$,
Bethany Kemp$^{1}$,
Andrew Badger$^{1}$,
Aurora Schmidt$^{1}$,
Krishna Murthy Jatavallabhula$^{3}$, \\
Celso M de Melo$^{4}$,
Lalithkumar Seenivasan$^2$,
Mathias Unberath$^2$,
Rama Chellappa$^2$,
\\
$^{1}$JHU APL,
$^{2}$JHU,
$^{3}$MIT,
$^{4}$DEVCOM ARL,
\thanks{$\dagger$Project Lead  *Equal Contribution}
}

\begin{document}

\maketitle
\thispagestyle{empty}
\pagestyle{empty}

\begin{abstract}

Robotic planning and execution in open-world environments is a complex problem due to the vast state spaces and high variability of task embodiment. Recent advances in perception algorithms, combined with Large Language Models (LLMs) for planning, offer promising solutions to these challenges, as the common sense reasoning capabilities of LLMs provide a strong heuristic for efficiently searching the action space. However, prior work fails to address the possibility of hallucinations from LLMs, which results in failures to execute the planned actions largely due to logical fallacies at high- or low-levels. To contend with automation failure due to such hallucinations, we introduce \textbf{ConceptAgent}, a natural language-driven robotic platform designed for task execution in unstructured environments. With a focus on scalability and reliability of LLM-based planning in complex state and action spaces, we present innovations designed to limit these shortcomings, including 1) Predicate Grounding to prevent and recover from infeasible actions, and 2) an embodied version of LLM-guided Monte Carlo Tree Search with self reflection. ConceptAgent combines these planning enhancements with dynamic language aligned 3d scene graphs, and large multi-modal pretrained models to perceive, localize, and interact with its environment, enabling reliable task completion. In simulation experiments, ConceptAgent achieved a 19\% task completion rate across three room layouts and 30 easy level embodied tasks outperforming other state-of-the-art LLM-driven reasoning baselines that scored 10.26\% and 8.11\% on the same benchmark. Additionally, ablation studies on moderate to hard embodied tasks revealed a 20\% increase in task completion from the baseline agent to the fully enhanced ConceptAgent, highlighting the individual and combined contributions of Predicate Grounding and LLM-guided Tree Search to enable more robust automation in complex state and action spaces. Additionally, in real-world mobile manipulation trials, conducted in randomized, low-clutter environments, a ConceptAgent-driven Spot robot achieved a 40\% task completion rate, demonstrating the performance of our perception system in real-world scenarios.
\end{abstract}

\section{INTRODUCTION}
The aspiration to have competent robotic counterparts that can reason about the world in an open manner and be commanded with natural language has been a long term goal of the robotics community. Recent advancements leveraging data-driven methodologies and robust robot models have showcased notable strides \cite{bring_robots_home,do_as_i_can,hand_eye_coord,learn_to_grasp}. However, existing systems often exhibit fragility, rigidity, and struggle when confronted with unforeseen circumstances. Many of these systems rely on being given a prescanned environment which limits their use to well known environments \cite{conceptgraphs,okrobot}.  Even the most expansive robotic models typically function optimally only within familiar settings \cite{rt1,rt2}. This fragility is particularly evident in scenarios with limited available robotic data, such as unstructured domestic environments. This lack of adaptability stands in stark contrast to the remarkable capabilities demonstrated by large-scale vision models, which excel in tasks such as semantic comprehension, object detection, and bridging visual and linguistic representations \cite{languagevision,openvit,detecting20k,okvqa, flamingo,dino,sam,conceptgraphs}.

\begin{figure*}[htbp]
    \centering
    \captionsetup{width=2\columnwidth}
    \includegraphics[width=1.6\columnwidth]{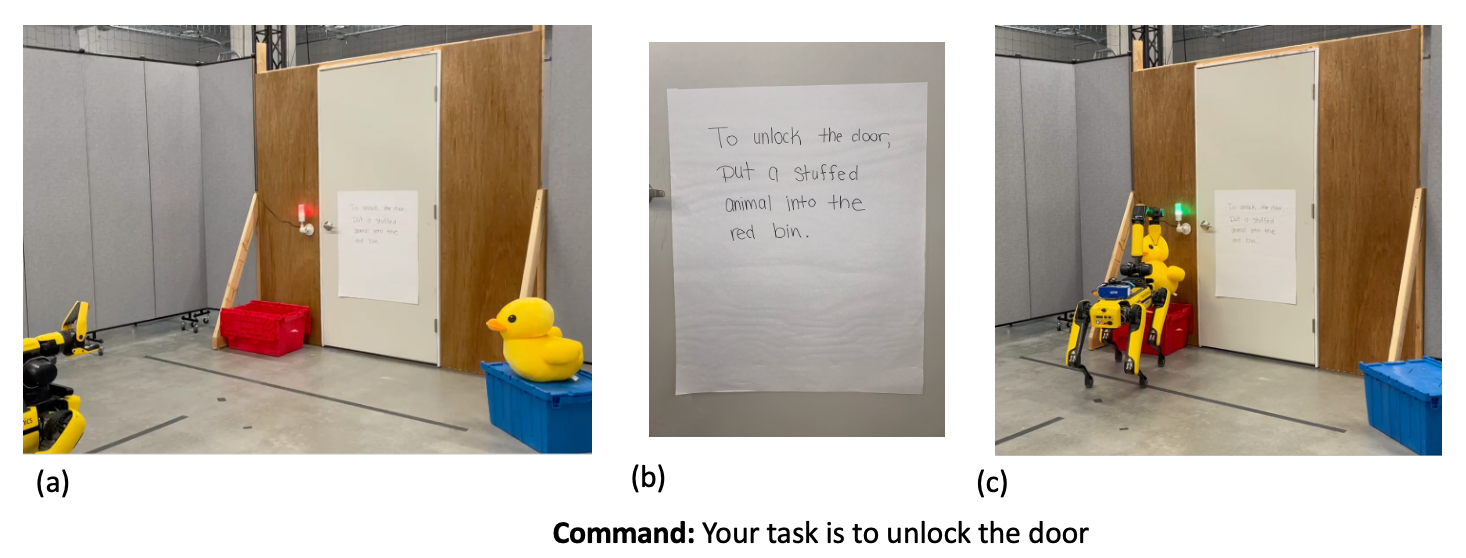}
    \caption{\textbf{ConceptAgent enables robust real-time, natural language-driven task execution in open-world environments.}  The problem requires the robot to operate in unfamiliar settings and manipulate novel objects to complete tasks described in unconstrained natural language. In this escape room motivated example, the task given is to "unlock the door".  The robot must not only identify objects but (a) understand the context of the scene including (b) a hand written note on the door with additional instructions.  (c) The ConceptAgent-driven Spot robot then proceeds to complete the task successfully without intervention. }
    \label{fig:unlock}
\end{figure*}

Despite the maturity of fundamental robotic skills like navigation, grasping, and object manipulation \cite{dex,anygrasp,graspnet,graspnet1b}, the integration of modern vision models with these foundational capabilities often yields underwhelming results. This challenge was underscored by the NeurIPS 2023 OVMM challenge, where the winning solution achieved a success rate of only 33\% \cite{ovmm1,ovmm2}. The complexity arises from a multitude of factors rather than a singular obstacle. Successful completion of general, long-horizon, embodied tasks requires intelligently generating a sequence of discrete actions, each of which are non-trivial both in their implementation and their execution. Each of these actions has the potential to accumulate and propagate error. Additionally, to create an actionable plan and recover when that plan goes awry requires a competent abstraction of the physical environment as well as a planner capable of fully leveraging that abstraction. Addressing these challenges necessitates a nuanced framework that seamlessly integrates natural language comprehension, full scene abstraction and understanding, and resilient reasoning.

In response to these challenges, we introduce \textbf{ConceptAgent}, a natural language-driven robotic platform designed for task execution in unstructured settings. ConceptAgent integrates advanced perception, manipulation, movement, speech, coding, and search primitives, allowing it to tackle a wide variety of tasks described in natural language. At its core, ConceptAgent combines large multi-modal pretrained models with novel planning and reasoning enhancements, enabling robust task execution in complex, dynamic environments.

Key innovations of ConceptAgent include: \begin{itemize} \item Formal Precondition Verification: ConceptAgent integrates precondition grounding, a mechanism that formally verifies action constraints before execution, preventing infeasible actions and facilitating failure recovery. This ensures the agent can maintain task progress, even in unstructured environments. \item LLM-Guided Monte Carlo Tree Search (LLM-MCTS): ConceptAgent employs LLM-guided tree search with self-reflection, enabling the agent to explore future states and refine action sequences dynamically. This approach significantly improves planning efficiency and task completion rates, even in large, open-world state spaces. \end{itemize}

To rigorously evaluate ConceptAgent, we conducted a series of experiments in both simulated and real-world environments. In simulation, we used the AI2Thor embodied agent simulator \cite{ai2thor} to benchmark performance across a diverse set of tasks and environments, allowing for the controlled assessment of our approach. Additionally, we tested ConceptAgent in real-world mobile manipulation tasks, where the robot navigated unfamiliar environments and interacted with novel objects. In a collection of 30 trials in randomized low clutter environments, we observed that the ConceptAgent-driven Spot robot completed \%40 of tasks successfully.

Our experiments yielded the following key findings: 
\begin{itemize} 
\item Precondition Grounding: Formal precondition verification significantly enhances the agent’s ability to recover from failure, leading to higher task completion rates and more reliable execution. 
\item LLM-Guided Tree Search and Reflection: By integrating LLM-guided tree search with closed-loop task execution and self-critique, ConceptAgent effectively refines its action sequences, achieving competitive task completion performance with far fewer expansion steps than traditional LLM-based systems. 
\end{itemize}

\section{Related Work}

\textbf{Traditional Robotic Task Planning } -  Task planning in robotics involves finding a sequence of actions to achieve specific goals within an environment. Traditional methods make use of domain-specific languages like PDDL \cite{pddl,pddl2} and finite state machines \cite{okrobot}, along with grammars and semantic parsing \cite{parsing}, search techniques \cite{search}, and heuristics \cite{huristics} to find solutions. These approaches can struggle with scalability and adaptability to real-world scenarios due to their incredibly large state-action space. Hierarchical, imitation, and reinforcement learning-based alternatives face challenges related to data demands and scalability \cite{rl,imitation}. Our approach leverages the common sense capabilities of Large Language Models (LLMs) to sidestep the brute force search problem associated with generating task plans in unconstrained environments. Leveraging realtime incremental 3D scene graphs for grounding \cite{conceptgraphs} and large pretrained multi-modal models \cite{spie}, our approach can formulate execution paths that were not preconceived, resulting in a capability to adapt and use affordances in ways that were not explicitly encoded by humans. 


\textbf{Task Planning with Large Language Models} - Task planning with LLMs, specifically translating natural language instructions into robotic task plans, is gaining traction in the field. Previous studies have effectively employed pre-trained LLMs' contextual learning abilities to generate actionable plans for embodied agents \cite{do_as_i_can,llmppddl,llmplanning,monologue,react,socratic,llmplanner}. However, a significant challenge remains in grounding these plans within the operational context of the robot. Past efforts make use of object detection models \cite{llmplanning}, PDDL environment representations \cite{llmppddl,sayplan}, or value functions \cite{do_as_i_can} for this purpose, but they are primarily effective in single-room environments and struggle with scalability. In our work, we improve scalability by re-imagining ConceptGraphs \cite{conceptgraphs} as a service with limited scope.  This allows the agent to select \textbf{only} the portions of the graph that are \textbf{most relevant} to a given task.

\textbf{Retrieval Augmented Generation for Task Planning} - Integrating external knowledge into LLMs has emerged as a promising approach to enhance their reliability. This involves using external tools to provide feedback or additional information to guide LLM output generation. Leveraging external knowledge in tool-based agents requires API calls to external tools \cite{check,toolformer} or textual feedback from the operating environment \cite{llmsim,monologue,errors}. Building on these concepts, we introduce an agent-based framework that reacts and replans based on the current observations to complete tasks specified in natural language.  

\section{ConceptAgent}

\begin{figure*}[h!]
    \centering
    \captionsetup{width=2\columnwidth}
    \includegraphics[width=1.6\columnwidth]{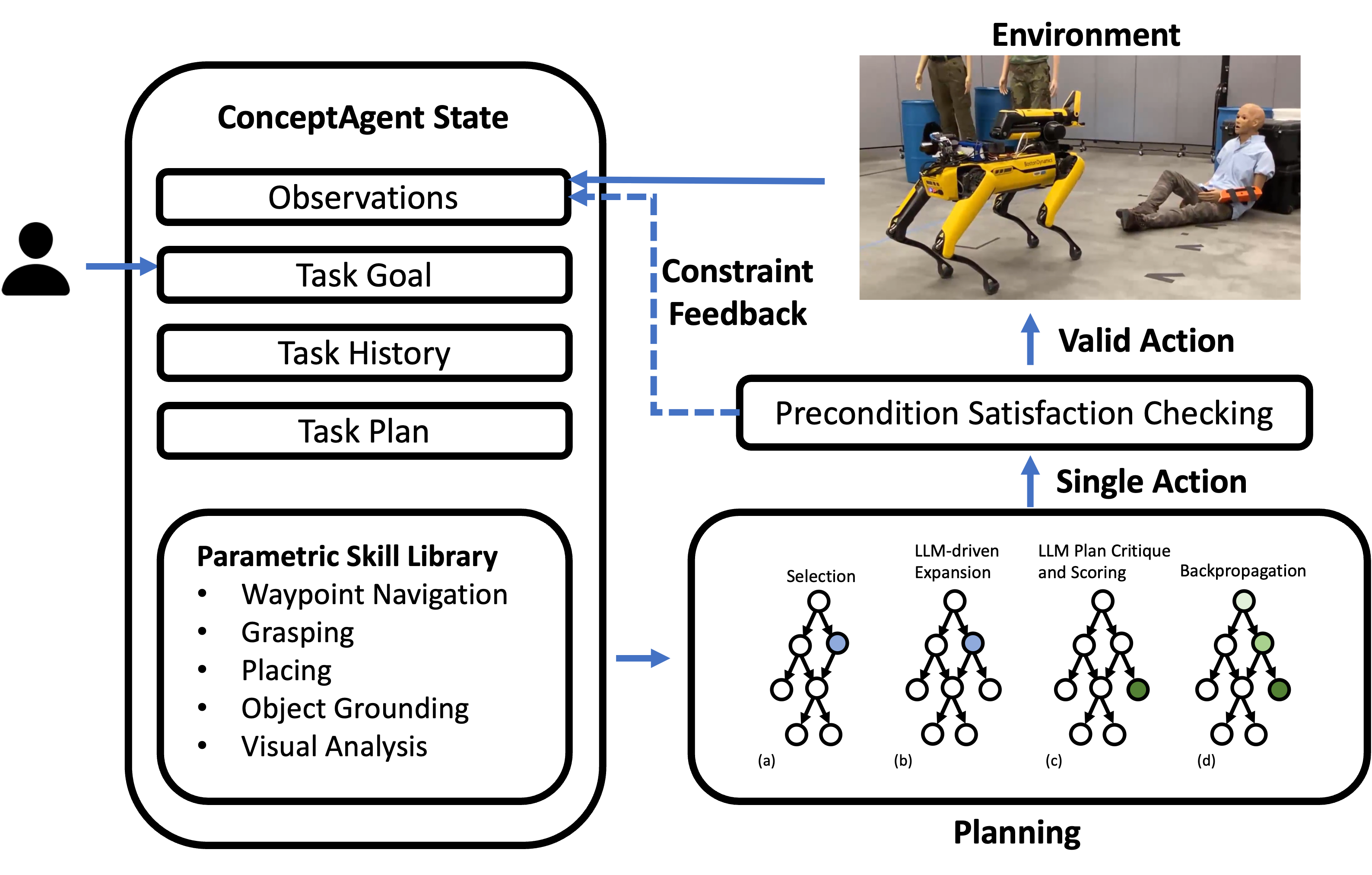}
    \caption{Overview of ConceptAgent closed loop task planning and execution.  State is composed of text description of the objective, task relevant observations, and task history.  That is combined with the details of a parametric skills library.  Tree-based planning is complemented with (b) LLM-based expansion and (c) LLM based critique and scoring. (a) Selection and (d) backpropagation are conducted like Monte-Carlo Tree Search.}
    \label{fig:conceptagent}
\end{figure*}

ConceptAgent enables natural language-driven task planning and execution. Figure \ref{fig:conceptagent} illustrates the framework.  The system responds to natural language task requests, plans and executes in a closed loop over a series of steps.  Intermediate feedback in the form of abstracted environment observations and information about unsatisfied action constraints (Precondition Grounding) allows the robot to adapt and recover in real-time.

\subsection{Problem Formulation}
We address the problem of real-time task execution in open-world environments driven by unconstrained natural language goals. In these scenarios, the key challenge lies in enabling a robotic system to understand and act upon complex, context-dependent language instructions, while simultaneously adapting to dynamic and unstructured environments.

The system must continuously interpret natural language expressions without predefined constraints, translating them into actionable plans in real-time. Full scene understanding is achieved through a combination of an incrementally-updating language-aligned 3d scene graph \cite{conceptgraphs} and large pretrained multi-modal models.

Given the open-world nature of the problem, the system must demonstrate the ability to generalize across varying environments and manage unknown objects, unforeseen obstacles, and unexpected environmental changes. The primary metric for success is task completion rate, underscoring the importance of reliable, autonomous decision-making in diverse and unpredictable contexts.

\subsection{LLM-guided Tree Search with Self-Reflection}

Inspired by \cite{llm-mcts}, ConceptAgent agent extends prior work in LLM-guided tree search to the real world domain. LLM-guided tree search follows the standard phases of traditional MCTS: \textit{selection}, \textit{expansion}, \textit{simulation}, and \textit{backpropagation}, but introduces LLM-based enhancements to expansion and simulation. Below, we detail each phase, emphasizing how the LLM guides planning and decision-making.

\subsubsection{Selection Based on Upper Confidence Bound}

The \textit{selection} phase is based on the Upper Confidence Bound (UCB1) algorithm \cite{uct-mcts}, which balances exploration and exploitation in selecting the next node to explore. For a node $s_t$ with child nodes $a_i$, the UCB1 criterion is used to select the action that maximizes the following equation:
\[
a^* = \arg\max_{a_i} \left[ Q(s_t, a_i) + c \cdot \sqrt{\frac{\log N(s_t)}{N(s_t, a_i)}} \right]
\]
where $Q(s_t, a_i)$ is the expected reward of taking action $a_i$ at state $s_t$, $N(s_t)$ is the number of times state $s_t$ has been visited, $N(s_t, a_i)$ is the number of times action $a_i$ has been taken from $s_t$, and $c$ is a constant controlling the exploration-exploitation trade-off. 

\subsubsection{LLM-Guided Expansion}

In the \textit{expansion} phase, traditional MCTS adds new child nodes representing possible future states. For embodied tasks, the state-action space explodes at even small depths in the tree search, requiring a powerful heuristic to filter out irrelevant states \cite{llm-mcts}. In ConceptAgent, we use the common sense of LLMs to serve as this heuristic. Given a current state $s_t$ and a natural language task goal $g$, the LLM generates a set of plausible actions $A_t = \mathcal{L}(s_t, g)$. These actions are then added to the tree as new nodes, representing potential transitions from $s_t$. This LLM-guided expansion allows the agent to explore a broader range of actions that align with the natural language task, thus avoiding state space explosion and improving task understanding and goal relevance.

\subsubsection{LLM-Based Critique for Simulation and Scoring}

In the traditional MCTS framework, the \textit{simulation} phase evaluates state transitions through random rollouts in the environment that backpropogate the value of the terminal state of that random rollout. For ConceptAgent, we replace this with a \textit{LLM-based critique} mechanism that assesses a planned sequence of actions holistically. At the leaf node of the search tree, rather than simulating state transitions, the LLM critic is given the sequence of actions planned from time $t$ to $t+k$. The LLM evaluates the efficiency, relevance, and goal alignment of the entire plan, producing a planning score that serves as the reward signal for backpropagation.

Formally, let $\tau = \{a_t, a_{t+1}, \dots, a_{t+k}\}$ be the sequence of planned actions from state $s_t$ over $k$ time steps. The LLM critic, $\mathcal{L}$, is tasked with evaluating the quality of this action sequence in achieving the task goal $g$. The critique score $C(\tau, g)$ reflects how well the action sequence satisfies the task requirements and the efficiency of the plan. The score incorporates penalties for unnecessary steps or inefficient actions.





\subsubsection{Backpropagation}

Once the critique score $C(s_{t+k}, g)$ is obtained, the \textit{backpropagation} phase proceeds as in standard MCTS. The critique score is propagated back through the tree, updating the value estimates $Q(s, a)$ for each state-action pair along the trajectory $\tau$. The update rule for each node is as follows:
\[
Q(s_t, a_i) \leftarrow \frac{1}{N(s_t, a_i)} \sum_{j=1}^{N(s_t, a_i)} C(\tau_j, g),
\]
where $N(s_t, a_i)$ is the number of times the action $a_i$ has been selected from state $s_t$, and $C(\tau_j, g)$ is the critique score from the $j$-th simulation. This ensures that the agent’s planning decisions reflect the accumulated knowledge from both successful and unsuccessful action sequences.

\subsection{Precondition Grounding}
Once MCTS with LLM-guided expansion and critique-based scoring identifies the optimal action sequence, ConceptAgent attempts to execute the chosen next action. If the tool execution fails, an agent must contextualize the failure with respect to the planning history to determine proper recovery measures. For example, upon failing to search inside a cabinet, the agent must analyze its previous actions and determine a reason for the execution failure. In this case, there are many possible reasons: the cabinet is not open, the agent has not moved within reach of the cabinet, the agent is currently holding an object and cannot execute the search tool due to manipulator constraints, etc. Through experiments, we find the baseline agent struggles to extrapolate such broad conclusions from the action history. Therefore, to improve agent recovery capabilities, we introduce Precondition Grounding as an enhancement to our LLM planner. The precondition grounding is autonomously generated, promoting scalability, and enables formal validation and recovery feedback during tool execution, greatly improving task completion.
\subsubsection{LLM-Derived Preconditions for Action Feasibility}
Precondition Grounding requires a defined set of preconditions for each tool. At inference time, these preconditions are verified by formal methods to ensure that a chosen tool can be executed given the perceived state of the environment. The creation of such preconditions generally lacks scalability \cite{pddl, pddl2}, but recent work has explored the ability of LLMs for precondition generation \cite{subbarao_llm_building_world_model}. Extending this idea, we prompt the agent to generate predicate based preconditions for each tool.

Formally, these preconditions encode the logical requirements for successful tool execution. Let $A = \{a_1, ..., a_n\}$ represent the set of $n$ actions available to the agent. Then, $P_i = LLM(S, n_i, d_i, atts)$ is the set of preconditions for $a_i$ generated by conditioning the LLM on the system prompt, $S$, action name, $n_i$, action description, $d_i$, and list of object and agent boolean attributes, $atts$. At time, $t$, we define the Formal Verification function, $F(s_t, P_c)$, as:
\[
F(s_t, P_c) = 
\begin{cases} 
1, & \text{if } P_c \text{ is satisfied in } s_t \\
0, & \text{if } P_c \text{ is not satisfied in } s_t
\end{cases}
\]
Where $c$ represents the index of the chosen action in $A$ and $s_t$ represents the state of the system at time, $t$. If $F(s_t, P_c) = 1$, then the action, $a_c$, is executed. If $F(s_t, P_c) = 0$, then the action is deemed infeasible and execution is aborted.

\subsubsection{Feedback for Future Planning}

When execution is aborted due to unmet preconditions, the system provides explicit feedback to the LLM agent, specifying which preconditions were unsatisfied. 

Let $U_c$ be the set of unsatisfied preconditions for action $c$ in $A$. If $F(s_t, P_c) = 0$, then:
\[
U_c = \{ p \in P_c \mid p \text{ is not satisfied in } s_t \}
\]
The set $U_c$ is formatted and returned to the LLM as feedback, allowing the agent to update its internal model of the environment and adjust future action sequences to avoid proposing actions whose preconditions are unlikely to be met. This iterative feedback loop refines the agent’s planning process, improving task completion rates over time by preventing the repetition of infeasible actions and guiding the agent towards valid recovery plans. Additionally, using feedback from $U_c$, the LLM revises its future plans, internalizing the state feedback to adjust the sequence of future actions. The agent can either attempt to satisfy the unsatisfied preconditions or generate alternative actions that are feasible given the current state $s_t$. This mechanism ensures that ConceptAgent's action plans are dynamically adapted to the environment.

\subsection{Implementation Details}
ConceptAgent make use of a library of parametric skills to complete tasks.  In physical experiments these skills include object localization, navigation, grasping, manipulation, visual question answering, speech, code execution, and web search.  Due to space constraints, we elaborate on these skills in the Appendix.

\section{Results}
The goals of our experiments were twofold: first, to evaluate the effectiveness of ConceptAgent’s planning and reasoning enhancements, including formal precondition verification and LLM-guided Monte Carlo Tree Search, in both simulated and real-world environments; and second, to assess its adaptability and task completion performance in diverse and unstructured scenarios, benchmarked against state-of-the-art generative AI reasoning approaches.
  
\subsection{Task Planning and Execution in Simulation}
Our simulation experiments were run in the AI2Thor simulator, a 3D simulator for kitchen environments that contains a variety of manipulable objects within the environment. To evaluate our planner, A set of 30 randomly constructed object rearrangement tasks were generated.  The object rearrangement tasks were characterized by simple, clear object rearrangement instructions e.g. ``The agent should pick up the credit card that is on the counter and place it on the kitchen drawer''.    An additional set of 40 tasks were hand crafted by a human expert. The tasks are split into 20 moderate tasks and 20 hard tasks. Moderate tasks are characterized by simple, object rearrangement instructions where task objects may be concealed within other objects e.g. ``Put a Tomato in a Cabinet.''.  The tomato may initially be in the refrigerator requiring additional exploration and manipulation.  Difficult tasks involve longer horizon goal states and increased ambiguity in the task statement e.g. ``Chill the tomato and put the bowl away in the cabinet.''. 


We first evaluate our agent's ability to autonomously generate preconditions for each provided action. To do this, a human expert first generated a set of ground truth preconditions for each possible agent action. We then compare the generated preconditions for each tool to that of the ground truth preconditions. The ground truth set contained 42 preconditions spanning across 10 different tools while the LLM generated set contained 38. We compared each precondition in the generated set to the ground truth set and noted 37 out of the 38 generated preconditions correctly matched to the 42 ground truth preconditions. We observe 6 ground truth preconditions were missing from the generated set. This represents an accuracy (correct generated / total generated) of 97.4\% and a recall (correct generated / total ground truth) of 88.1\%. Although 6 ground truth preconditions were missing from the generated set, this will be no worse than the baseline agent without any verification mechanism. In short, only 2.6\% of the LLM generated preconditions are incorrect and thus have real potential to cause harm to planning beyond that of the baseline agents without enhancements. We further evaluate the LLM generated preconditions empirically through their impact on planning performance in Table \ref{tab:sim_experiments}.


\subsection{Comparison to Baselines}

We compared our approach to several state-of-the-art closed-loop generative AI-based reasoning baselines adapted for embodied tasks. The approaches were evaluated with 30 object rearrangement benchmark tasks across 3 room layouts. Specifically, we evaluated the performance of these reasoning approaches alongside multiple levels of expansion in ConceptAgent. As a positive control, we included results from a baseline approach that utilized an LLM nearly 10 times larger than those used in other methods.

Our results, summarized in Table \ref{tab:baselines}, show that the baseline methods, including ReAct \cite{react} and Tree of Thoughts (ToT) \cite{tot}, both using an 8-billion-parameter LLM, achieved task completion rates of 10\% and 8\%, respectively. In comparison, ConceptAgent, with 10 and 20 expansions, achieved task completion rates of 14\% and 19\%, respectively. These results highlight the effectiveness of our approach, demonstrating a clear improvement in task performance with additional ConceptAgent expansions.

Notably, with 20 expansions, ConceptAgent's performance (19\% task completion) approached that of the ReAct baseline, which relied on an LLM nearly 10 times larger. This illustrates the efficiency of ConceptAgent in leveraging smaller models while still achieving competitive results through improved planning and reasoning mechanisms.

\begin{figure*}[h!]
    \centering
    \captionsetup{width=2\columnwidth}
    \includegraphics[width=1.6\columnwidth]{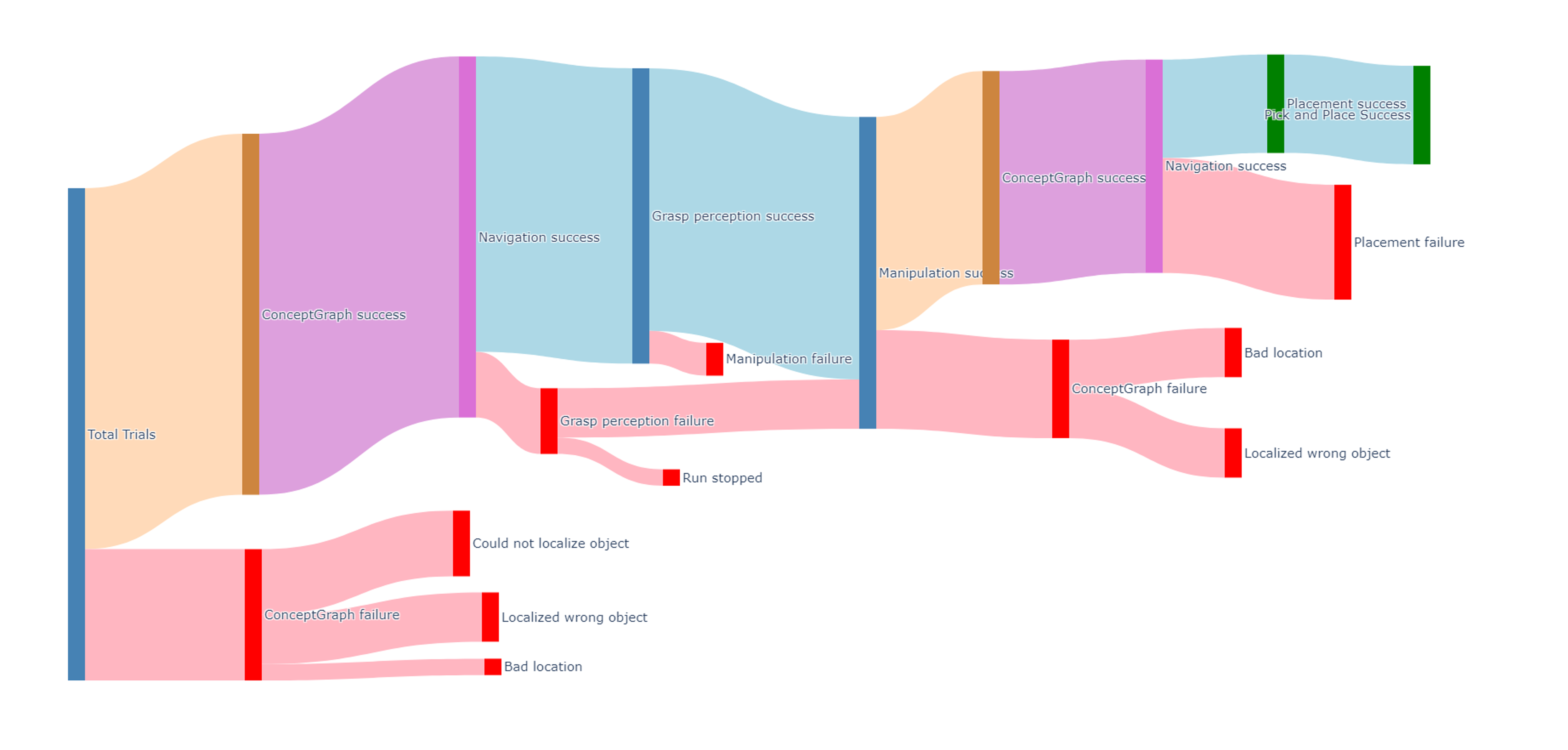}
    \caption{\textbf{Evaluation of Physical Mobile Manipulation for Open Vocabulary Object Rearrangement }  - Object rearrangement success and failure cases aggregated over all levels of clutter, broken down by mode of failure / success. From left to right, we show the performance of the system to locate the object for object rearrangement, navigate to it, perceive it, grasp it, locate the destination, navigate to the destination, and place the object into the receptacle.}
    \label{fig:pick-and-place-sankey}
\end{figure*}

\subsection{Ablation Experiments}
Next, we explored ablations of ConceptAgent on the moderate and hard benchmarks. In these experiments we ablate the two key planning enhancements: Precondition Grounding and LLM-MCTS. The results shown in Table \ref{tab:sim_experiments} report task success rate separately for the moderate and hard task benchmarks along with an overall average over the 40 tasks.

Without predicte grounding or LLM-driven MCTS, ConceptAgent completes 5\% of the moderate and 5\% of the hard tasks.  When combined with precondition grounding, we find that ConceptAgent was able to complete 15\% of the moderate tasks and 10\% of the hard tasks successfully. This shows that, while Llama3.1 70B tends to choose tools under invalid execution constraints, it is able to recover from these planning errors given the appropriate feedback in the form of unsatisfied preconditions. LLM-MCTS was evaluated with 5 expansion steps in the ablation experiments.  When ConceptAgent includes LLM-driven MCTS and precondition grounding we observed the highest task completion on both the moderate and hard benchmark tasks achieving 35\% and 15\% respectively.

\begin{table}
\centering
\setlength{\tabcolsep}{4pt}
\begin{tabular}{lccccc|c|} \hline
\rule{0pt}{2.2ex} Method & LLM & Moderate & Hard & Overall\\ \hline\hline
\rule{0pt}{2.2ex} Base Agent (BA) & 70b & 5\% & 5\% & 5\% \\ 
\rule{0pt}{2.2ex} BA + LLM-MCTS & 70b & 10\% & 0\% & 5\% \\ 
\rule{0pt}{2.2ex} BA + PG & 70b & 15\% & 10\% & 12.5\% \\ \
\rule{0pt}{2.2ex}CA (BA + LLM-MCTS + PG) & 70b & \textbf{35\%} & \textbf{15}\% & \textbf{25}\% \\ \hline
\end{tabular}
\caption{Ablations of ConceptAgent (CA) on Moderate and Hard benchmark tasks. We report success rate for each benchmark and an overall average.  In these experiments, we ablate precondition grounding (PG) and generative tree search planning (LLM-MCTS) from the ReAct baseline. LLama 3.1 70b is used for all experiments. }
\label{tab:sim_experiments}
\end{table}

\begin{table}
\setlength{\tabcolsep}{2pt}
\centering
\begin{tabular}{lccc} \hline
\rule{0pt}{2.2ex}Method & Expansions & LLM & Task Completion \\ \hline\hline
\rule{0pt}{2.2ex}ReAct \cite{react} & N/A &llama3.1 8b & 10.26\% \\ 
\rule{0pt}{2.2ex}ReAct+ToT \cite{tot} & N/A &llama3.1 8b & 8.11\% \\ 
\rule{0pt}{2.2ex}ConceptAgent [Ours] & 10 &llama3.1 8b & 13.89\% \\ 
\rule{0pt}{2.2ex}ConceptAgent [Ours] & 20 &llama3.1 8b & 18.92\% \\ 
\rule{0pt}{2.2ex}ReAct \cite{react} & N/A &llama3.1 70b & 22.5\% \\ \hline
\end{tabular}
\caption{Comparison to embodied reasoning baselines on the easy benchmark of object rearrangement tasks across multiple floorplans. We compared ConceptAgent to several state-of-the-art closed loop LLM-based reasoning baselines adapted for embodied tasks.}
\label{tab:baselines}
\end{table}

\begin{table}[]
\setlength{\tabcolsep}{4pt}
\begin{tabular}{lccccccc}
\hline
\rule{0pt}{2.2ex}         & Locate & Move & Percept. & Manip. & Locate & Move & Place \\ \hline\hline
\rule{0pt}{2.2ex}Success  & 22     & 22    & 18       & 19     & 13     & 13    & 6     \\ 
\rule{0pt}{2.2ex}Failures & 8      & 0     & 4        & 2      & 6      & 0     & 7     \\ 
\rule{0pt}{2.2ex}Total    & 30     & 22    & 22       & 21     & 19     & 13    & 13    \\ 
\rule{0pt}{2.2ex}SSR      & 73\%   & 100\% & 81\%     & 90\%   & 68\%   & 100\% & 40\%  \\ \hline

\end{tabular}
\caption{Step-wise success rates for skills in object rearrangement tasks.}

\vspace{-2mm}
\label{tab:stepwise}
\end{table}

\begin{table}[]
\begin{tabular}{lccccccc}
\hline
\rule{0pt}{2.2ex}Clutter      & Locate & Percept. & Manip. & Locate  & Place & Overall \\ \hline\hline
\rule{0pt}{2.2ex}High  & 60\%     & 66\%     & 83\%   & 80\%    & 25\% & 10\%     \\ 
\rule{0pt}{2.2ex}Medium & 80\%      & 88\%   & 85\%   & 33\%    & 50\% & 10\%     \\ 
\rule{0pt}{2.2ex}Low    & 80\%     & 88\%   & 100\%   & 88\%    & 57\% & 40\%  \\ 
\rule{0pt}{2.2ex}Average      & 73\%   & 81\%     & 90\%   & 68\%    & 46\% & 20\% \\ \hline

\end{tabular}
\caption{Effects of environment clutter - Three different levels of clutter were considered with 10 tasks selected at random.  The table reports step-wise success rates as well as overall task success rate.  Reasoning and navigation was not included in the table for brevity because they were 100\% successful in these trials. }
\vspace{-2mm}
\label{tab:clutter}
\end{table}

\subsection{Physical Trial for Object Rearrangement }

In these trials, we aimed to characterize failure modes in physical mobile manipulation experiments.  To benchmark the system's performance, we began with object rearrangement tasks, modeled after of experiments of prior work \cite{okrobot,uniteam,come-robot,spot-compose}. These tasks are formulated as "pick up object X and place it on/in object Y". We ran 30 iterations of these tasks, selecting objects and their destinations at random. Every 5 iterations, the scene layout would be randomized, and every 10 iterations, the scene would be decluttered like prior work \cite{okrobot}, where for each declutter procedure, we select the select 1/3 of the objects that showed difficulty for the system in terms of semantic ambiguity or grasping difficulty and relocate the remaining objects to accessible locations.  In these trials, ConceptAgent was evaluated with GPT-4, single expansion, and standard feedback. \textbf{Metrics} -- To evaluate these results, we recorded the following metrics: Success rate, Partial success rate / step-wise success rate (SSR).

Although the open vocabulary step-wise success rates (SSR) were relatively high (Table \ref{tab:stepwise}), completing the task requires successfully executing multiple sequential steps. As such, the overall task success rate is the product of the step-wise success rates, resulting in a total success rate of 20\%. This occurred despite the system generating highly accurate plans in each trial and achieving 100\% successful navigation.

Table \ref{tab:clutter} presents the overall task success rates under different levels of clutter. After decluttering, we observed an increase in success rate from 10\% to 40\%, highlighting that both visual clutter and semantic ambiguity are critical factors affecting task performance. Figure \ref{fig:pick-and-place-sankey} illustrates the step-wise and overall success rates in a Sankey diagram, mapping the primary failure modes observed during the trials.

\section{Conclusions}
Large multi-modal pretrained models are essential for enhancing the robustness of robotic plan execution in open-world environments. These models enable ConceptAgent to integrate perception and reasoning, allowing for more accurate and adaptable task execution in diverse, unstructured scenarios.

\section{Acknowledgements}
This research was sponsored by the Army Research Laboratory and was accomplished under Cooperative Agreement Number W911NF-21-2-0211. The views and conclusions contained in this document are those of the authors and should not be interpreted as representing the official policies, either expressed or implied, of the Army Research Office or the U.S. Government. The U.S. Government is authorized to reproduce and distribute reprints for Government purposes notwithstanding any copyright notation herein. DISTRIBUTION A. Approved for public release; distribution unlimited.

\bibliographystyle{IEEEtran}
\bibliography{refs}

\section*{Appendix}
\definecolor{codeblue}{rgb}{0.38039216, 0.61568627, .8       }
\definecolor{codepink}{rgb}{1.        , 0.41568627, 0.83529412}
\definecolor{codegray}{rgb}{0.5,0.5,0.5}
\definecolor{codepurple}{rgb}{0.58,0,0.82}
\definecolor{backcolour}{rgb}{0.98,0.98,0.98}

\lstdefinestyle{mystyle}{
    backgroundcolor=\color{backcolour},   
    commentstyle=\color{codeblue},
    keywordstyle=\color{codepink},
    numberstyle=\tiny\color{codegray},
    stringstyle=\color{codepurple},
    basicstyle=\ttfamily\scriptsize,
    breakatwhitespace=false,         
    breaklines=true,                 
    captionpos=b,                    
    keepspaces=true,                 
    showspaces=false,                
    showstringspaces=false,
    showtabs=false,                  
    tabsize=2
}
\lstset{style=mystyle}

\lstdefinelanguage{gptprompt}{
  basicstyle=\ttfamily\footnotesize,
  keywords = {action_information
  },
  keywordstyle=\color{red},
  keywords = [2]{User, LLM-Planner},
  keywordstyle = [2]\color{blue},
}

\section{System prompt for generating preconditions}
\label{app:sec:generating_preconditions}
We give our agent a brief description of the task of generating preconditions using the PDDL format. PDDL was chosen as this is a commonly used domain language for generating predicate based preconditions and effects, thus it was natural to ask the LLM agent in this format as the agent has likely seen PDDL format extensively in its training data.
\begin{lstlisting}[language=gptprompt, caption=Precondition Grounding System Prompt - Part 1]
You are to generate definitions for a robotic
action listed below such that the preconditions 
required to perform the action successfully and 
effects that occur when the action is performed 
successfully are appropriately described. This 
should be specified in PDDL format using
predicates to specify the preconditions and 
effects.

The action information will be provided in the following JSON format:
{{"Action Name": str, "Action Description": str}}
\end{lstlisting}
We continue the prompt by providing the agent with a list of boolean predicates that it can choose from. These predicates are a list of object and agent attributes that represent the state of each object and the agent in the environment. The agent must choose between these predicates when defining preconditions for actions.
\begin{lstlisting}[language=gptprompt, caption=Precondition Grounding System Prompt - Part 2]
New predicates can be created, but they MUST 
inherit from the starting predicates below.

(moveable ?x - objectId): true if the object is moveable
(breakable ?x - objectId): true if the object is breakable
(canFillWithLiquid ?x - objectId): true if the object can be filled with liquid
(isToggled ?x - objectId): true if the object is toggled on
(pickupable ?x - objectId): true if the object is able to be picked up
(isOpen ?x - objectId): true if the object is open
(isBroken ?x - objectId): true if the object is broken
(visible ?x - objectId): true if the object is visible
(receptacle ?x - objectId): true if the object is a receptacle
(openable ?x - objectId): true if the object is able to be opened
(isPickedUp ?x - objectId): true if the object is picked up
(toggleable ?x - objectId): true if the object is able to be toggled
(isFilledWithLiquid ?x - objectId): true if the object is filled with liquid
(cookable ?x - objectId): true if the object is able to be cooked
(isCooked ?x - objectId): true if the object is cooked
(isWaterSource ?x - objectId): true if the object is a water source
(isHoldingObject): true if the agent is holding an object
\end{lstlisting}
We continue the prompt by providing the agent with a description of 3 logical modifiers: `and', `exists', and `when'. The `and' modifier is true if and only if each precondition nested inside it is satisfied, the `exists' modifier is true if there is at least one object in the environment for which all of the nested preconditions are satisfied, and the `when' modifier is true if 1) the first condition in the modifier is False 2) the first condition in the modifier is true and the second condition in the modifier is also true. This is essentially an if/then modifier stating: if the first condition is true, then the second condition must also be true. We find that these three modifiers allow for a diverse set of preconditions that can satisfy a variety of complex environmental constraints for each action.
\begin{lstlisting}[language=gptprompt, caption=Precondition Grounding System Prompt - Part 3]
To combine predicates for preconditions and 
effects, you can use 3 different modifiers:

Modifier 1:
name: and
description: useful for if you want to combine predicates.
example: an object satisfies the following if it is openable and not toggleable.
(and 
    (openable ?x)
    (not (toggleable ?x))
)

Modifier 2:
name: when
description: useful for if you want to require a predicate conditionally based on another predicate.
example: if an object is openable, it will satisfy the following if it is not open. if the object is not openable, it will always satisfy the below exampl.
(when 
    (openable ?x)
    (not (isOpen ?x))    
)

Modifier 3:
name: exists
description: useful for if you do not know a specific object id, but want to make sure there is an object in the environment that satisfies certain predicates.
example: the below is satisfied there exists an object in the environment that is held and isCooked.
(exists (?y)
    (and 
        (isCooked ?y)
        (isPickedUp ?y)
    )

)
\end{lstlisting}
The system prompt terminates with an example set of precondition generation task for one of the agent tools before prompting the agent to `Begin!'. We repeat this prompt for each action in the set of actions provided to the agent where the action\_information is a parameter to the prompt providing the action name and description.
\begin{lstlisting}[language=gptprompt, caption=Precondition Grounding System Prompt - Part 4]
Below is an example task. Please follow the below format starting from "Here is a list of rules for each of the calls that you have provided." in your response:

Please generate the precondition and effect definition for the following
based on the list of recorded state transitions.

Action Information:
{{ "Action Name": "Search Object", "Action Description": "Return a list of all items inside or on a target object specified by object id. Input format in JSON: {{'object_id': 'Cabinet_4'}}."}}

Parameters:
1. ?x - object: the target object to search

Begin!

Here is the definition outlining the preconditions and effects of the action.
Preconditions:
```pddl
(and
    (receptacle ?x)
    (visible ?x)
    (when
        (and 
            (openable ?x)
        )
        (and
            (isOpen ?x)
        )
    )
    (not (isHoldingObject))
)
```
Justification:  
It makes sense that in order to search an object, that object likely must first
be visible. Also, in order to search an object like a cabinet or a refrigerator,
the object must first be open so that you can see the inside. But for other objects,
like a countertop, the object is not openable, so the isOpen predicate should only
apply to those objects that are openable. Also, in order to search an object I 
must have a free hand, so I should not be holding anything. For the above reasons, 
the choice of preconditions is justified.

Finished!

Please generate the precondition and effect definition for the following
based on the list of recorded state transitions.

Action Information:
{action_information}

Parameters:
1. ?x - object: the target object to search

Begin!
\end{lstlisting}
\section{Full list of LLM generated preconditions}
\label{app:sec:preconditions}
\lstset{
    escapeinside={(*@}{@*)},          
}
\begin{lstlisting}
(*@\textcolor{green}{correct}@*): in the ground truth, generated by the LLM
(*@\textcolor{blue}{missing}@*): in the ground truth, not generated by the LLM
(*@\textcolor{red}{incorrect}@*): not in the ground truth, generated by the LLM

Pick Pick Up Object
 preconditions
  and
   (*@\color{green}{pickupable = true}@*)
   (*@\color{green}{visible = true}@*)
   (*@\color{green}{isPickedUp = false}@*)
   (*@\color{green}{isHoldingObject = false}@*)

Place Object
 preconditions
  and
   (*@\color{green}{isHoldingObject = true}@*)
   (*@\color{green}{receptacle = true}@*)
   (*@\color{green}{visible = true}@*)
   if
    (*@\color{green}{openable = true}@*)
   then
    (*@\color{green}{isOpen = true}@*)

Open Object
 preconditions
  and
   (*@\color{green}{openable = true}@*)
   (*@\color{green}{isOpen = false}@*)
   (*@\color{green}{visible = true}@*)
   (*@\color{red}{isBroken = false}@*)
   (*@\color{blue}{isHoldingObject = false}@*)

Close Object
 preconditions
  and
   (*@\color{green}{openable = true}@*)
   (*@\color{green}{isOpen = true}@*)
   (*@\color{green}{visible = true}@*)
   (*@\color{blue}{isHoldingObject = false}@*)

Toggle Object On
 preconditions
  and
   (*@\color{green}{toggleable = true}@*)
   (*@\color{green}{isToggled = false}@*)
   (*@\color{green}{visible = true}@*)
   (*@\color{blue}{isHoldingObject = false}@*)

Toggle Object Off
 preconditions
  and
   (*@\color{green}{toggleable = true}@*)
   (*@\color{green}{isToggled = true}@*)
   (*@\color{blue}{visible = true}@*)
   (*@\color{blue}{isHoldingObject = false}@*)

Search Object
 preconditions
  and
   (*@\color{green}{receptacle = true}@*)
   (*@\color{green}{visible = true}@*)
   (*@\color{green}{isHoldingObject = false}@*)
   if 
    (*@\color{green}{openable = true}@*)
   then
    (*@\color{green}{isOpen = true}@*)

Fill Held Object With Water
 preconditions
  and
   (*@\color{green}{isHoldingObject = true}@*)
   exists
    and
     (*@\color{green}{isPickedUp = true}@*)
     (*@\color{green}{canFillWithLiquid = true}@*)
     (*@\color{green}{isFilledWithLiquid = false}@*)
   exists
    and
     (*@\color{green}{isWaterSource = true}@*)
     (*@\color{green}{isToggled = true}@*)
     (*@\color{blue}{visible = true}@*)

Pour Water Into
 preconditions
  and
   (*@\color{green}{isHoldingObject = true}@*)
   (*@\color{green}{canFillWithLiquid = true}@*)
   (*@\color{green}{visible = true}@*)
   exists
    and
     (*@\color{green}{isPickedUp = true}@*)
     (*@\color{green}{isFilledWithLiquid = true}@*)


\end{lstlisting}


\section{Task examples}

Below we provide example tasks from each of the 3 subsets (Easy, Medium, Hard).

\textbf{Easy Subset Examples:}
\begin{enumerate}
    \item \textit{The agent should pick up the dish sponge and place it in the bowl that is on the counter top.}
    \item \textit{The agent should pick up the bread that is on the counter top and place it in the microwave.}
    \item \textit{The agent should pick up the bowl that is on the counter top and place it in the sink basin.}
\end{enumerate}

\textbf{Medium Subset Examples:}
\begin{enumerate}
    \item \textit{Fill a vase with water.}
    \item \textit{Place a fork in a cup.}
    \item \textit{Find the spoon in the drawer and place it in the sink.}
\end{enumerate}

\textbf{Hard Subset Examples:}
\begin{enumerate}
    \item \textit{Cook an egg.}
    \item \textit{Water the plant.}
    \item \textit{Cook a potato without using the stove. Leave the potato inside what it was cooked in.}
\end{enumerate}

    \begin{figure*}[h!]
    \centering
    \captionsetup{width=2\columnwidth}
    \includegraphics[width=1.6\columnwidth]{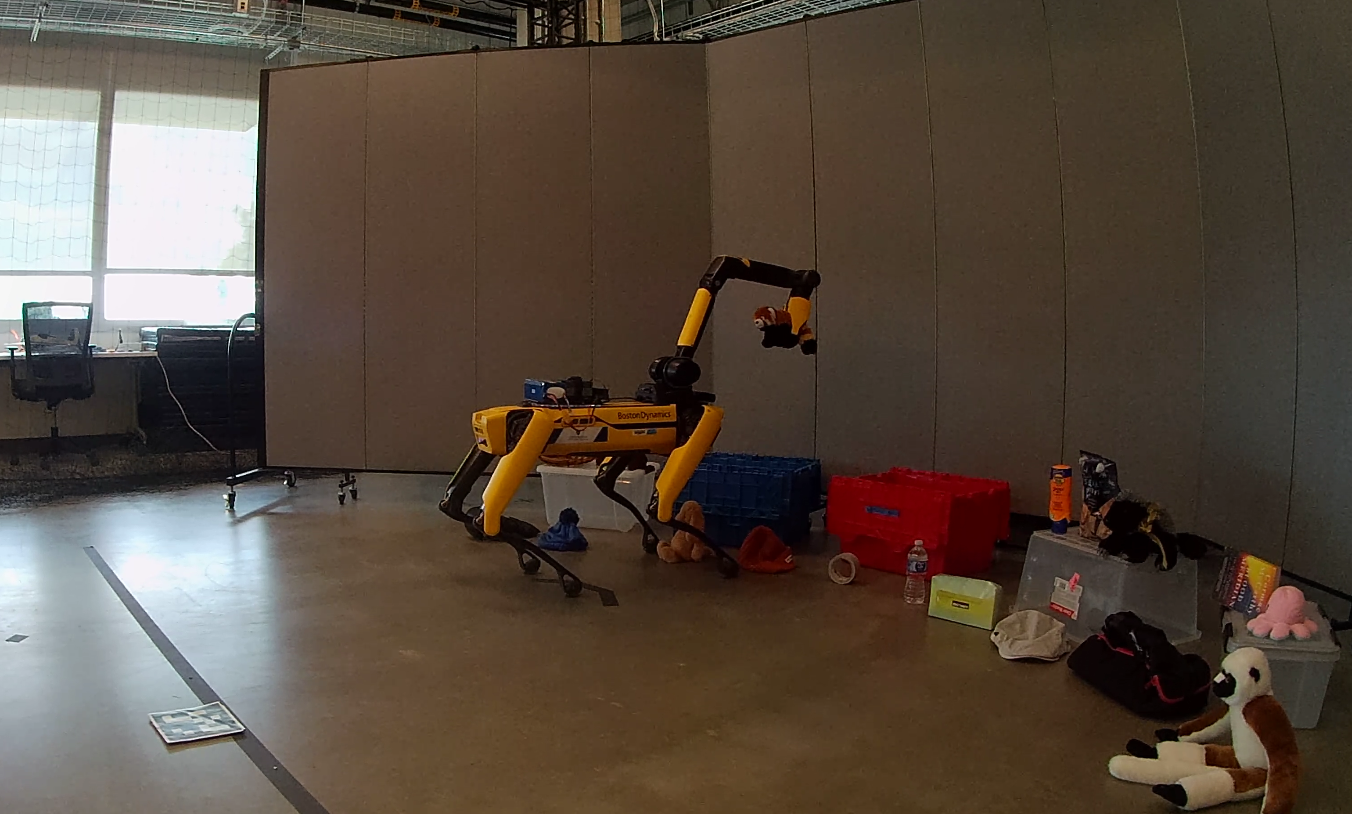}
    \caption{\textbf{Mobile Manipulation Trials} - The trials were aimed at categorizing failure modes for physical mobile manipulation.}
    \label{fig:card}
\end{figure*}
\section{Example Successful Execution of a Hard task.}
\label{app:sec:successful_hard}

\begin{figure*}[h!]
    \centering
    \captionsetup{width=2\columnwidth}
    \includegraphics[width=1.6\columnwidth]{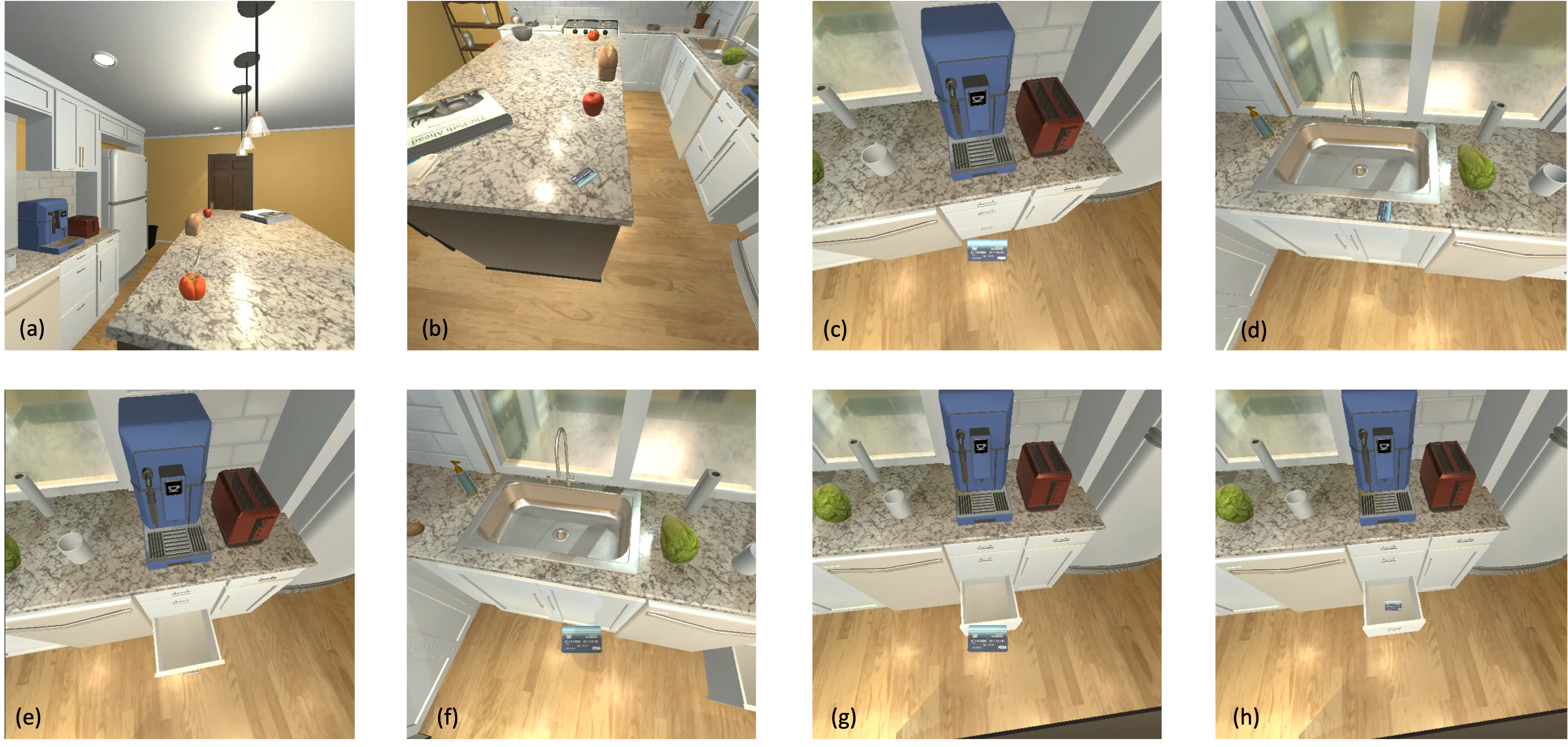}
    \caption{\textbf{Example Task Completion in AI2Thor by ConceptAgent} \emph{Task: Put the credit card on the counter into the kitchen drawer.} (a) the agent starts in the kitchen, (b) after some exploration the agent finds the credit card, (c) the agent takes the credit card to the kitchen drawer, but it's closed, (d) the agent adapts by placing the card by the kitchen sink, before (e) moving back to the drawer and opening it.  (f) the agent moves back to the sink to pick up the card again, before (g) moving back to the open drawer, and (h) placing the credit card in the drawer to complete the task.}
    \label{app:fig:card}
\end{figure*}

\lstdefinelanguage{tool_plan}{
  basicstyle=\ttfamily\footnotesize,
  keywords = {Thought
  },
  keywordstyle=\color{blue},
  keywords = [2]{User, LLM-Planner},
  keywordstyle = [2]\color{blue},
}

\lstset{
    language=tool_plan,
    escapeinside={(*@}{@*)},          
}

\begin{figure*}[t]
\begin{lstlisting}[caption=Successful  for a Difficult task (1/3).]
Task: Clear off the dining room table.
Thought: To clear off the dining room table, I need to identify all objects currently on the table and then pick them up one by one. First, I'll randomly explore to find the dining room table and other objects.
Action: Randomly Explore
Action Input: {'input': None}
Observation: Tool Completed Successfully: New objects discovered!
Thought: I have discovered new objects in the environment. I should now get the list of all discovered objects to identify the dining room table and the items on it.
Action: Get Discovered Objects
Action Input: {'input': None}
Observation: Tool Completed Successfully: Here is a list of the object ID's of all of the discovered objects: ['Apple_1', 'Blinds_1', 'Bread_1', 'ButterKnife_1', 'Cabinet_1', 'Cabinet_2', 'Cabinet_3', 'Cabinet_4', 'Cabinet_5', 'Cabinet_6', 'Cabinet_7', 'Cabinet_8', 'CoffeeMachine_1', 'CounterTop_1', 'CounterTop_2', 'DiningTable_1', 'Drawer_1', 'Drawer_2', 'Drawer_3', 'Faucet_1', 'Floor_1', 'GarbageCan_1', 'Lettuce_1', 'Microwave_1', 'Mug_1', 'Pan_1', 'Plate_1', 'Pot_1', 'SinkBasin_1', 'Sink_1', 'SoapBottle_1', 'Spatula_1', 'Stool_1', 'StoveBurner_1', 'StoveBurner_2', 'StoveBurner_3', 'StoveBurner_4', 'StoveKnob_1', 'StoveKnob_2', 'StoveKnob_3', 'StoveKnob_4', 'Toaster_1', 'Tomato_1', 'Window_1', 'WineBottle_1']
Thought: The dining table has been identified as 'DiningTable_1'. I need to inspect it to find out what items are currently on it.
Action: Inspect Object
Action Input: {'object_id': 'DiningTable_1'}
Observation: Tool Completed Successfully: The queried object is currently inside or on top of the following objects: ['Floor|-01.72|-00.08|+02.08'].
The quried objects currently controls the following objects: None.
Thought: It seems that the dining table is positioned on the floor, but the inspection did not return any items on top of it. I'll need to search the dining table to find out if there are any objects placed on it.
Action: Search Object
Action Input: {'object_id': 'DiningTable_1'}
Observation: Tool Completed Successfully: Here is a list of objects found inside or on the target receptacle: ['Knife_1', 'SaltShaker_1', 'Bread_1', 'PepperShaker_1', 'Lettuce_1', 'Potato_1']
Thought: I have identified several objects on the dining table: 'Knife_1', 'SaltShaker_1', 'Bread_1', 'PepperShaker_1', 'Lettuce_1', and 'Potato_1'. I will now proceed to pick up each object one by one.
Action: Pick Up Object
Action Input: {'object_id': 'Knife_1'}
Observation: Tool Completed Successfully: Target object was picked up!
Thought: I have successfully picked up the knife from the dining table. I will now proceed to place it somewhere else, such as the countertop.
Action: Place Object
Action Input: {'object_id': 'CounterTop_1'}
Observation: Tool Failed: the target object is not visible, 
Thought: It seems that the countertop is not currently in my line of sight. I need to adjust my position so that I can place the knife on the countertop.
Action: Adjust Positioning
Action Input: {'object_id': 'CounterTop_1'}
Observation: Tool Completed Successfully: Readjusted position relative to target object!
Thought: My positioning has been adjusted relative to the countertop. I will now try to place the knife on the countertop again.
Action: Place Object
Action Input: {'object_id': 'CounterTop_1'}
Observation: Tool Completed Successfully: Held object was successfully placed in target object!
Thought: I have successfully placed the knife on the countertop. Next, I will proceed to pick up the remaining objects on the dining table: 'SaltShaker_1', 'Bread_1', 'PepperShaker_1', 'Lettuce_1', and 'Potato_1', and move them to an appropriate location, such as the countertop.
Action: Pick Up Object
Action Input: {'object_id': 'SaltShaker_1'}
Observation: Tool Failed: the target object is not visible, 
Thought: The salt shaker is not currently in my line of sight. I need to adjust my position so that I can see and pick up the salt shaker from the dining table.
Action: Adjust Positioning
Action Input: {'object_id': 'SaltShaker_1'}
Observation: Tool Completed Successfully: Readjusted position relative to target object!
Thought: My positioning has been adjusted relative to the salt shaker. I will now try to pick up the salt shaker from the dining table.
Action: Pick Up Object
Action Input: {'object_id': 'SaltShaker_1'}
Observation: Tool Completed Successfully: Target object was picked up!
Thought: I have successfully picked up the salt shaker from the dining table. I will now proceed to place it on the countertop.
Action: Place Object
Action Input: {'object_id': 'CounterTop_1'}
Observation: Tool Failed: the target object is not visible, 
\end{lstlisting}
\end{figure*}

\begin{figure*}[t]
\begin{lstlisting}[caption=Successful Execution of a Difficult task (2/3).]
Thought: It seems that the countertop is not currently in my line of sight again. I need to adjust my position so that I can place the salt shaker on the countertop.
Action: Adjust Positioning
Action Input: {'object_id': 'CounterTop_1'}
Observation: Tool Completed Successfully: Readjusted position relative to target object!
Thought: My positioning has been adjusted relative to the countertop. I will now try to place the salt shaker on the countertop again.
Action: Place Object
Action Input: {'object_id': 'CounterTop_1'}
Observation: Tool Completed Successfully: Held object was successfully placed in target object!
Thought: I have successfully placed the salt shaker on the countertop. Next, I will proceed to pick up the remaining objects on the dining table: 'Bread_1', 'PepperShaker_1', 'Lettuce_1', and 'Potato_1', and move them to an appropriate location, such as the countertop.
Action: Pick Up Object
Action Input: {'object_id': 'Bread_1'}
Observation: Tool Failed: the target object is not visible, 
Thought: The bread is not currently in my line of sight. I need to adjust my position so that I can see and pick up the bread from the dining table.
Action: Adjust Positioning
Action Input: {'object_id': 'Bread_1'}
Observation: Tool Completed Successfully: Readjusted position relative to target object!
Thought: My positioning has been adjusted relative to the bread. I will now try to pick up the bread from the dining table.
Action: Pick Up Object
Action Input: {'object_id': 'Bread_1'}
Observation: Tool Completed Successfully: Target object was picked up!
Thought: I have successfully picked up the bread from the dining table. I will now proceed to place it on the countertop.
Action: Place Object
Action Input: {'object_id': 'CounterTop_1'}
Observation: Tool Failed: the target object is not visible, 
Thought: The countertop is not in my line of sight again. I need to adjust my position to place the bread on the countertop.
Action: Adjust Positioning
Action Input: {'object_id': 'CounterTop_1'}
Observation: Tool Completed Successfully: Readjusted position relative to target object!
Thought: My positioning has been adjusted relative to the countertop. I will now try to place the bread on the countertop again.
Action: Place Object
Action Input: {'object_id': 'CounterTop_1'}
Observation: Tool Completed Successfully: Held object was successfully placed in target object!
Thought: I have successfully placed the bread on the countertop. Next, I will proceed to pick up the remaining objects on the dining table: 'PepperShaker_1', 'Lettuce_1', and 'Potato_1', and move them to an appropriate location, such as the countertop.
Action: Pick Up Object
Action Input: {'object_id': 'PepperShaker_1'}
Observation: Tool Failed: the target object is not visible, 
Thought: The pepper shaker is not currently in my line of sight. I need to adjust my position so that I can see and pick up the pepper shaker from the dining table.
Action: Adjust Positioning
Action Input: {'object_id': 'PepperShaker_1'}
Observation: Tool Completed Successfully: Readjusted position relative to target object!
Thought: My positioning has been adjusted relative to the pepper shaker. I will now try to pick up the pepper shaker from the dining table.
Action: Pick Up Object
Action Input: {'object_id': 'PepperShaker_1'}
Observation: Tool Completed Successfully: Target object was picked up!
Thought: I have successfully picked up the pepper shaker from the dining table. I will now proceed to place it on the countertop.
Action: Place Object
Action Input: {'object_id': 'CounterTop_1'}
Observation: Tool Failed: the target object is not visible, 
Thought: The countertop is not in my line of sight again. I need to adjust my position to place the pepper shaker on the countertop.
Action: Adjust Positioning
Action Input: {'object_id': 'CounterTop_1'}
Observation: Tool Completed Successfully: Readjusted position relative to target object!
Thought: My positioning has been adjusted relative to the countertop. I will now try to place the pepper shaker on the countertop again.
Action: Place Object
Action Input: {'object_id': 'CounterTop_1'}
Observation: Tool Completed Successfully: Held object was successfully placed in target object!
\end{lstlisting}
\end{figure*}

\begin{figure*}[t]
\begin{lstlisting}[caption=Successful Execution of a Difficult task including Predicate Grounding (3/3).]
Thought: I have successfully placed the pepper shaker on the countertop. Next, I will proceed to pick up the remaining objects on the dining table: 'Lettuce_1' and 'Potato_1', and move them to the countertop.
Action: Pick Up Object
Action Input: {'object_id': 'Lettuce_1'}
Observation: Tool Failed: the target object is not visible, 
Thought: The lettuce is not currently in my line of sight. I need to adjust my position so that I can see and pick up the lettuce from the dining table.
Action: Adjust Positioning
Action Input: {'object_id': 'Lettuce_1'}
Observation: Tool Completed Successfully: Readjusted position relative to target object!
Thought: My positioning has been adjusted relative to the lettuce. I will now try to pick up the lettuce from the dining table.
Action: Pick Up Object
Action Input: {'object_id': 'Lettuce_1'}
Observation: Tool Completed Successfully: Target object was picked up!
Thought: I have successfully picked up the lettuce from the dining table. I will now proceed to place it on the countertop.
Action: Place Object
Action Input: {'object_id': 'CounterTop_1'}
Observation: Tool Failed: the target object is not visible, 
Thought: The countertop is not in my line of sight again. I need to adjust my position to place the lettuce on the countertop.
Action: Adjust Positioning
Action Input: {'object_id': 'CounterTop_1'}
Observation: Tool Completed Successfully: Readjusted position relative to target object!
Thought: My positioning has been adjusted relative to the countertop. I will now try to place the lettuce on the countertop again.
Action: Place Object
Action Input: {'object_id': 'CounterTop_1'}
Observation: Tool Failed: the target object is not visible, 
Thought: My repeated attempts to place objects on the countertop have failed because the countertop is not consistently in my line of sight. Instead, I'll try placing the objects in a different location that is more consistently visible, such as the floor.
Action: Place Object
Action Input: {'object_id': 'Floor_1'}
Observation: Tool Completed Successfully: Held object was successfully placed in target object!
Thought: I have successfully placed the lettuce on the floor. Now, I need to pick up the last remaining object on the dining table, which is the potato, and move it to the floor.
Action: Pick Up Object
Action Input: {'object_id': 'Potato_1'}
Observation: Tool Completed Successfully: Target object was picked up!
Thought: I have successfully picked up the potato from the dining table. I will now proceed to place it on the floor.
Action: Place Object
Action Input: {'object_id': 'Floor_1'}
Observation: Tool Completed Successfully: Held object was successfully placed in target object!
Thought: I have successfully cleared all the objects from the dining room table and placed them in different locations, either on the countertop or on the floor. The task of clearing off the dining room table is now complete.
Final Answer: The dining room table has been successfully cleared of all objects.
\end{lstlisting}
\end{figure*}
\section{LLM Reflection for MCTS Plan Evaluation}
\label{app:sec:critique}
LLM-based reflection replaces simulation and evaluation in the typical MCTS algorithm.  The LLM-based reflection provides a third party evaluation of the plan.  The LLm is prompted to score the quality of the plan and provide justification given the goal and the steps of the plan.

\begin{lstlisting}[language=gptprompt, caption=Self-critique and Reflection for Plan Evalution]
Score the quality the plan and provide a justification

Strictly use the following format:

Goal: The stated goal
Plan: the actions to take, should make use of {{"Tool Names": str}}
Justification: Justification for the score
Score: single number from 1-10 indicating the quality of the plan

Begin! 

Goal: {{"goal": str}}
Plan: {{"plan steps": str}}
\end{lstlisting}

\section{LLM-driven Action Candidate Expansion}
\label{app:sec:expansion}
LLM-based action expansion is used to propose candidate actions and parameterization given the current state of the environment and the stated goal.

\begin{lstlisting}[language=gptprompt, caption=Action Candidate Expansion]
You are a robot who can use tools to take action in the real world.  You have access to the following tools:

{tools}

Strictly use the following format:

Question: the task you must complete
Thought: you should always think about what to do and replan if needed
Action: the action to take, should be one of [{tool_names}]
Action Input: the input to the action
Observation: the result of the action
... (this Thought/Action/Action Input/Observation can repeat N times)
Thought: I now know the final answer
Final Answer: Verify the completion of the task

Begin! 

Question: {input}
{agent_scratchpad}
\end{lstlisting}
\section{Implementation Details for Robotic Skills}
\label{app:sec:skills}

\subsection{Real-time 3d Scene Graphs}
The perception system maintains an abstraction of the physical environment through a dynamic 3d scene graph that updates in real-time. Inspired by \cite{conceptgraphs}, our perception pipeline leverages several off-the-shelf computer vision models to accurately segment, embed, and project objects into a 3 dimensional pointcloud where open vocabulary object retrieval can be performed. A significant shortcoming of prior work required pre-scanning of the environment and offline analysis before queries could be executed. This severely limits the ability of an embodied agent, as exploration and discovery of the environment are critical to the completion of open ended, general tasks. To circumvent this limitation, our approach introduces a fast combination of Segment Anything (SAM) (everything mode) and CLIP for instance segmentation and image embedding in real-time. The bounding boxes of candidate objects are then used as crops which are then passed to SAM to extract only the pixels in the image related to the candidate object. Next, CLIP is used to get an embedding of the segmented candidate object which is then stored to support natural language based grounding to object locations and non-axis aligned 3d bounding boxes. We use CLIP embeddings instead of discrete object classes to enable a retrieval that is robust to nuances in the natural language query (e.g. cup vs mug) by matching input queries to candidate objects through cosine similarity between embeddings.

\subsection{Waypoint Navigation}
In this work, we implement waypoint navigation using the ROS2 Nav2 system, supported by real-time point clouds generated from a forward-facing stereoscopic ZED camera. The ZED camera captures depth information through stereo vision, producing 3D point clouds that provide detailed spatial perception of the environment.

The Nav2 system utilizes these point clouds for obstacle detection and localization, integrating them into the map for dynamic path planning. The robot follows a set of predefined waypoints, adjusting its path as obstacles are detected or the environment changes. The stereoscopic point clouds enhance the navigation precision by improving obstacle avoidance and route planning, particularly in complex, unstructured environments.

The ROS2 Nav2 stack is configured to leverage the ZED camera's real-time data stream to build a local costmap, which continuously updates as the robot progresses along its route. This allows for both reactive and deliberative path planning, making the system suitable for navigation tasks that require high spatial accuracy.

\subsection{Manipulation}
For object manipulation, we leverage SAM \cite{sam} (Segment Anything Model) in "everything mode" and CLIP to reidentify the query object in the robot's gripper camera frame. SAM generates an image mask for the target object, and a grasp point is automatically selected within the identified mask. The robot then performs visual servoing, combined with inverse kinematics, to compute and achieve the precise 6-DoF grasp position and orientation using Boston Dynamics APIs. The manipulation pipeline is divided into separate skills for grasping and placing, ensuring robust execution of complex tasks involving object interaction in dynamic environments. This approach allows for accurate, real-time manipulation of objects with minimal manual intervention.

\subsection{Additional Skills}
\begin{itemize}
    \item \textbf{Code execution} -- Code execution was supported by PythonREPL
    \item \textbf{Web Search} - Web search was backed by SerpAPI
    \item \textbf{Speech Recognition and Synthesis} -- Nvidia Riva
\end{itemize}


\end{document}